\documentclass[pdflatex,sn-mathphys-num]{sn-jnl}

\usepackage{graphicx}%
\usepackage{multirow}%
\usepackage{amsmath,amssymb,amsfonts}%
\usepackage{amsthm}%
\usepackage{mathrsfs}%
\usepackage[title]{appendix}%
\usepackage{textcomp}%
\usepackage{manyfoot}%
\usepackage{booktabs}%
\usepackage{algorithm}%
\usepackage{algorithmicx}%
\usepackage{algpseudocode}%
\usepackage{listings}%
\usepackage{xcolor}

\usepackage[utf8]{inputenc}
\usepackage[T1]{fontenc}
\usepackage{tabularx}           
\usepackage{setspace}
\usepackage{ragged2e}
\usepackage{caption}
\usepackage{adjustbox}\usepackage{booktabs}\usepackage{multirow}\usepackage{caption}

\theoremstyle{thmstyleone}%
\theoremstyle{thmstyletwo}%

\theoremstyle{thmstylethree}%

\begin{document}

\title[Article Title]{A Scoping Review of Synthetic Data Generation by Language Models in Biomedical Research and Application: Data Utility and Quality Perspectives}








\author[1]{\fnm{Hanshu} \sur{Rao} \sfx{BS}}\email{hrao@memphis.edu}

\author[1]{\fnm{Weisi} \sur{Liu} \sfx{MS}}\email{wliu9@memphis.edu}

\author[2]{\fnm{Haohan} \sur{Wang} \sfx{PhD}}

\author[3]{\fnm{I-Chan} \sur{Huang} \sfx{PhD}}

\author[4]{\fnm{Zhe} \sur{He} \sfx{PhD}}

\author*[1]{\fnm{Xiaolei} \sur{Huang} \sfx{PhD}}\email{xiaolei.huang@memphis.edu}

\affil*[1]{\orgdiv{Department of Computer Science}, \orgname{University of Memphis}, \orgaddress{\city{Memphis}, \postcode{38152}, \state{TN}, \country{United States}}}

\affil[2]{\orgdiv{School of Information Sciences}, \orgname{University of Illinois Urbana-Champaign}, \orgaddress{\city{Champaign}, \postcode{61820}, \state{IL}, \country{United States}}}

\affil[3]{\orgdiv{Epidemiology and Cancer Control}, \orgname{St Jude Children's Research Hospital}, \orgaddress{\city{Memphis}, \postcode{38105}, \state{TN}, \country{United States}}}

\affil[4]{\orgdiv{School of Information}, \orgname{Florida State University}, \orgaddress{\city{Tallahassee}, \postcode{32306}, \state{FL}, \country{United States}}}


\abstract{
Synthetic data generation using large language models (LLMs) demonstrates substantial promise in addressing biomedical data challenges and shows increasing adoption in biomedical research. 
This study systematically reviews recent advances in synthetic data generation for biomedical applications and clinical research, focusing on how LLMs address data scarcity, utility, and quality issues with different modalities.
We conducted a scoping review following PRISMA-ScR guidelines and searched literature published between 2020 and 2025 through PubMed, ACM, Web of Science, and Google Scholar.
A total of 59 studies were included based on relevance to synthetic data generation in biomedical contexts.
Among the reviewed studies, the predominant data modalities were unstructured texts (78.0\%), tabular data (13.6\%), and multimodal sources (8.4\%). 
Common generation methods included LLM prompting (74.6\%), fine-tuning (20.3\%), and specialized models (5.1\%). 
Evaluations were heterogeneous: intrinsic metrics (27.1\%), human-in-the-loop assessments (44.1\%), and LLM-based evaluations (13.6\%). 
However, limitations and key barriers persist in data modalities, domain utility, resource and model accessibility, and standardized evaluation protocols.
Future efforts may focus on developing standardized, transparent evaluation frameworks and expanding accessibility to support effective applications in biomedical research.
}



\maketitle

\section{Introduction}\label{Introduction}



Data is the key to train reliable AI models for broad biomedical and clinical applications, such as medical diagnosis~\cite{liu2025time,jones2025examining}, therapeutic treatments~\cite{stade2024large}, and drug discovery~\cite{blanco2023role}. 
However, obtaining massive, privacy-preserving, and high-quality data faces critical challenges, such as data availability, noisy and missing data, and legal regulations. 
Increasing biomedical studies are turning attention to synthetic data generation, a process of creating artificial datasets that accurately replicate the statistical and structural properties of real-world data. Nonetheless, creating high-quality synthetic data remains challenging due to the inherent heterogeneity, complexity, and variability characteristic of biomedical data. 
Large language models (LLMs) may offer a promising solution by enhancing the availability and utility of synthetic datasets in critical applications, such as clinical note analysis for stroke thrombolysis contraindication identification~\cite{chen2025automated}, mental health diagnosis through automated interview assessment~\cite{wu2024callm}, and radiology report generation for fracture misdiagnosis detection~\cite{liu2025generating}. 

Synthetic data generation in biomedical and clinical fields has achieved significant advances by LLMs (e.g., GPT-4~\cite{achiam2023gpt} and Llama 3~\cite{dubey2024llama}) in recent years, as shown in Figure \ref{fig:Publication_trend} and \ref{fig:PRISMA-ScR_flow}.
For example, LLMs have been applied successfully to obtain clinical narratives and simulate real patient records of mental health for diagnosis and behavior analysis~\cite{han2024chain}, such as predicting suicide~\cite{ghanadian2024socially} and depression patterns~\cite{bucur2024leveraging}. 
The strong generation capability not only relieves data privacy concerns~\cite{sarkar2024identification} to facilitate model developments but also may promote data quality~\cite{xu-etal-2024-knowledge} to support clinical decision-making. 
For example, synthetic corpora have augmented classification models for cardiovascular and Alzheimer’s disease diagnosis~\cite{li2023two}; 
and a study develops a multi-agent dialogue generator (NoteChat) that creates synthetic patient-physician conversations to improve clinical documentation~\cite{wang2024notechat}.
Nevertheless, questions and uncertainties remain regarding the best practices, validation approaches, and specific application areas for those LLM-generated synthetic data.

The goal of this scoping review is to comprehensively summarize and assess recent research publications on the biomedical and clinical utilities leveraging LLMs for synthetic data generation and its quality evaluation. 
While multiple reviewing studies~\cite{figueira2022survey, long-etal-2024-llms, li-etal-2023-synthetic} have covered LLM-generated synthetic data for general domains, studies on biomedical and clinical domains remain underexplored. 
For example, a close study~\cite{li-etal-2023-synthetic} examines recent developments of generation algorithms and models on news and social media domains instead of biomedical fields. 
Several recent review studies~\cite{murtaza2023synthetic, gonzales2023synthetic, smolyak2024large, loni2025review} have examined the topic in health and clinical fields, however, their focus on methodologies and models often overlooks fundamental aspects of synthetic data itself, such as types, utility, accessibility, and data quality —– gaps that this study aims to address.
Specifically, through this study, we will identify and present the current state-of-the-art models, technical methods, evaluations or assessments of synthetic data quality, and gaps and opportunities for future research. 
We seek to answer a concrete yet unsolved question: \textit{what biomedical and clinical applications can be effectively addressed using LLM-generated synthetic data, and how?}

\section{Results}\label{Results}

\begin{figure}[ht]
\centering
\includegraphics[width=0.7\linewidth]{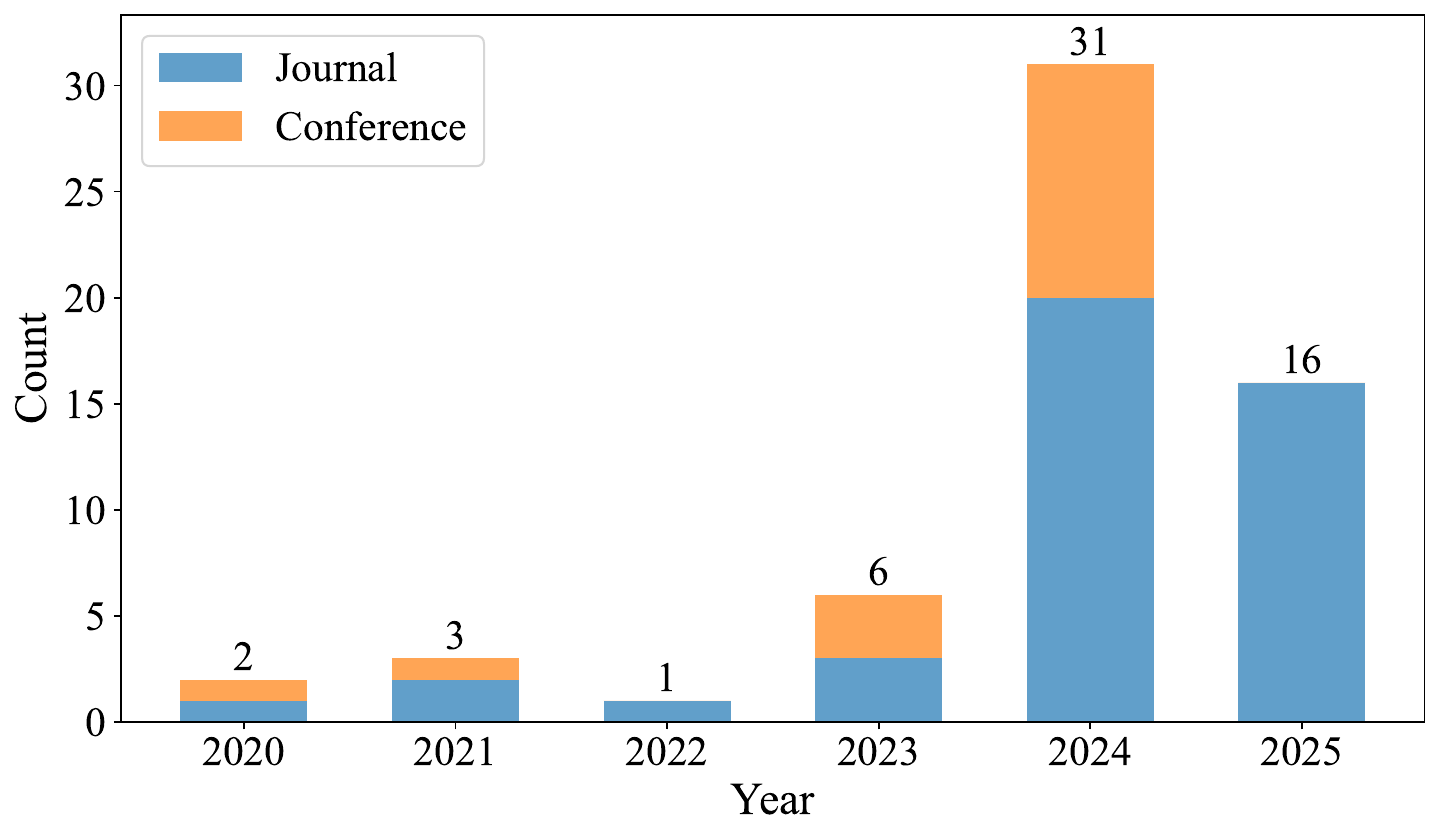}
\caption{
Publications between Jan 01, 2020 and April 05, 2025. Orange and blue colors refer to peer-reviewed conference and journal publications, respectively. 
We can observe a surging increase on the biomedical synthetic data related studies.}
\label{fig:Publication_trend}
\end{figure}

Our search and initial review resulted in 132 articles and finalized with 59 articles for this scoping review. XH, HW, and HR reviewed full texts of the articles and removed 73 articles due to ineligible article types ($n=2$), no large language model ($n=7$), unrelated to biomedical data generation ($n=18$), and not peer-reviewed and published in a journal or conference proceedings ($n=46$). 
Inter-rater reliability, assessed using Cohen’s $\kappa$~\cite{landis1977measurement}, indicated almost perfect agreement for both screening phases ($\kappa=0.96$ for title/abstract screening, $n=307$; $\kappa=0.95$ for full-text screening, $n=132$).
Fifty-nine articles met our criteria and were included in our subsequent analyses. 
In terms of publication venues and formats, the included literature consists of 43 journal articles (72.9\%) and 16 peer-reviewed conference papers (27.1\%), leaving 59 studies to be included in this scoping review. 
Figure \ref{fig:Publication_trend} shows that the study number of biomedical synthetic data generation climbs steadily over the years. 
Particularly, the substantial growth after 2022 reflects a particularly evolving trend of synthetic biomedical data generation via LLM, an emerging generation tool in recent years.

We present an overview of recent progress in synthetic biomedical data generation and application from three perspectives, aiming to provide readers with an insightful understanding of current practices and inform future research directions:

\begin{enumerate}
  \item Synthetic data types and medical applications: Research has explored broad synthetic data types of clinical text, tabular data, and multimodal information (e.g., images, audio, and sensor signals). 
  These data types cover a range of applications, including EHR analysis, medical imaging, telemedicine, mental health, and clinical trial matching, addressing challenges related to data scarcity and privacy restrictions.
  
  \item Synthetic data generation methodology: Evolving methodologies encompass prompting-based generation, knowledge infusion, and multi-agent and multimodal approaches. 
  These advancements enhance the semantic consistency, efficiency, and scalability of synthetic biomedical data generation.

  \item Quality assessments and evaluation metrics: Ensuring data quality and utility remains a priority. 
  Current studies employ evaluation strategies targeting fidelity, utility, and privacy protection. 
  Such strategies integrate statistical metrics, human-in-the-loop review, automated model assessments, and privacy testing to establish a robust framework for verifying the reliability and compliance of synthetic data.
\end{enumerate}

\begin{figure}[!htbp]

\captionsetup{type=table}
\centering

\captionof{table}{Summary of data types, generation methods, accessibility, purposes and medical applications in the collected studies. 
Languages include English (EN), Dutch (NL), French (FR), Chinese (ZH), and Arabic (AR). 
Generation methods comprise fine-tuning (SFT: supervised, DAFT: domain-adaptive, TAFT: task-adaptive, IT: instruction tuning, None) and prompting (ZS: zero-shot, FS: few-shot, INST: instruction-driven, CoT: reasoning-augmented, RAG: knowledge-augmented). 
Synthetic data purpose includes Training (used independently), Supplement (used with real data for joint training), and Privacy (used for privacy preservation such as sharing or de-identification). 
Data accessibility is classified as Yes (publicly available), Request (available upon request), and No (not publicly available or not specified).
Medical applications cover question answering (QA), information extraction (IE), and social determinants of health (SDoH).
}

\label{tab:generation_approach}

\begin{adjustbox}{width=\textwidth, totalheight=0.76\textheight}
\begin{tabular}{@{}lccccllcll@{}}
\multirow{2}{*}{\textbf{Study}} & 
\multirow{2}{*}{\textbf{Modality}} & 
\multirow{2}{*}{\textbf{Lang.}} & 
\multirow{2}{*}{\textbf{Year}} &
\multirow{2}{*}{\textbf{Generative Model}} & 
\multicolumn{2}{c}{\textbf{Approach}} & 
\multirow{2}{*}{\textbf{Accessible}} & 
\multirow{2}{*}{\textbf{Purpose}} & 
\multirow{2}{*}{\textbf{Medical Application}} \\
\cmidrule(lr){6-7}
& & & & & {\textbf{Fine-tuning}} & \textbf{Prompting} & & &\\
\midrule
 
\cite{shakeri-etal-2020-end} & Text & EN & 2020 & BART & TAFT & INST & No & Training & Biomedical QA \\ 
\cite{ive2020generation} & Text & EN & 2020 & Transformer & SFT & INST & No & Training & Phenotype inference \\ 
\cite{li2021synthetic} & Text & EN & 2021 & GPT-2 & TAFT & ZS & Yes & Training & Clinical IE \\ 
\cite{libbi2021generating} & Text & NL & 2021 & GPT-2 & DAFT & ZS & No & Training & De-identification \\ 
\cite{chintagunta-etal-2021-medically} & Text & EN & 2021 & GPT-3 & None & FS, RAG & No & Training & Dialogue   summarization \\ 
\cite{lu2022textual} & Text & EN & 2022 & GPT-2 & TAFT & ZS & No & Training & Readmission prediction \\ 
\cite{peng2023study} & Text & EN & 2023 & GPT-3 & None & ZS & Yes & Training & General \\ 
\cite{hiebel-etal-2023-synthetic} & Text & FR & 2023 & GPT-2, Bloom & DAFT & ZS & Yes & Training & Clinical IE \\ 
\cite{khademi2023data} & Text & EN & 2023 & GPT-2 & TAFT & ZS & No & Training & Syndromic detection \\ 
\cite{spitale2023exploring} & Text & EN & 2023 & GPT-3.5 & IT & INST & Yes & Training & Assisted suicide \\ 
\cite{theodorou2023synthetic} & Tabular & EN & 2023 & HALO & TAFT & INST & Yes & Training & General \\ 
\cite{li2023two} & Text & EN & 2023 & Llama, GPT-4 & None & CoT & Request* & Training & Alzheimer's disease \\ 
\cite{sufi2024addressing} & Text & EN & 2024 & ChatGPT API & None & INST & Yes & Training & Discharge summary \\ 
\cite{wu2024callm} & Text & EN & 2024 & GPT-4 & None & INST, CoT & Request* & Training & PTSD diagnosis \\ 
\cite{chen2024chatffa} & Image & ZH, EN & 2024 & GPT-3.5 & None & INST & No & Training & Medical imaging \\ 
\cite{litake2024constructing} & Text & EN & 2024 & GPT-3.5, Llama-2 & None & INST & Yes & Training & Acute renal failure \\ 
\cite{sarkar2024identification} & Text & EN & 2024 & GPT-3.5 & None & INST & No & Privacy & Phenotype inference \\ 
\cite{nievas2024distilling} & Text & EN & 2024 & GPT-4 & None & INST, CoT & No & Training & Cohort selection \\ 
\cite{wang2024dke} & Text & EN & 2024 & GPT-3.5 & None & INST, RAG & No & Supplement & Cohort selection \\ 
\cite{moser2024generating} & Text & DE & 2024 & Zephyr, GPT-4 & None & INST, RAG & No & Training & Emergency medicine \\ 
\cite{bird2024generative} & Text & EN & 2024 & Mistral & None & ZS, INST & Yes & Training & Psychological therapy \\ 
\cite{jeong2024improving} & Text & EN & 2024 & GPT-4, Llama-2 & IT & ZS, FS & Yes & Training & Medical QA \\ 
\cite{zafar2024ki-mag} & Text & EN & 2024 & T5 & TAFT & RAG & Request & Training & Medical QA \\ 
\cite{xu-etal-2024-knowledge} & Text & EN & 2024 & GPT-3.5 & None & FS, RAG & Yes & Training & General \\ 
\cite{guevara2024large} & Text & EN & 2024 & GPT-3.5/4 & None & FS, INST & Yes & Supplement & SDoH extraction \\ 
\cite{ehrett2024leveraging} & Tabular & EN & 2024 & ChatGPT & None & FS & No & Supplement & Medical education \\ 
\cite{wang2024notechat} & Text & EN & 2024 & GPT-3.5 & None & INST, CoT & Yes & Supplement & Dialogue generation \\ 
\cite{gabriel2024development} & Text & EN & 2024 & GPT-3.5 & None & ZS, INST & No & Training & SDoH extraction \\ 
\cite{gao2024pressinpose} & Tabular & EN & 2024 & GPT-4 & None & FS & No & Training & Gait rehabilitation \\ 
\cite{kweon2024publicly} & Text & EN & 2024 & GPT-3.5 & None & FS, INST & Yes & Training & General \\ 
\cite{weerasinghe2024real} & Audio & EN & 2024 & GPT-4 & None & INST & Yes & Supplement & EMS assistance \\ 
\cite{delmas2024relation} & Text & EN & 2024 & Vicuna & None & INST & Yes & Training & Clinical IE \\ 
\cite{ghanadian2024socially} & Text & EN & 2024 & GPT-3.5, Flan-T5, Llama-2 & None & ZS, FS & No & Supplement & Suicide prediction \\ 
\cite{zeinali2024symptom} & Text & EN & 2024 & GPT-4 & None & INST & No & Training & Cancer symptom extraction \\ 
\cite{mishra2024synthetic} & Text & EN & 2024 & ChatGPT & None & INST & Yes & Training & Clinical summary \\ 
\cite{zhang-etal-2024-synthetic} & Text & EN & 2024 & GPT-3.5 & None & INST, RAG & Yes & Training & Medical QA \\ 
\cite{dobhal2024synthetic} & Tabular & - & 2024 & GPT-4o & None & FS, CoT & No & Supplement & Nursing care \\ 
\cite{wang2024taxonomy} & Text & EN & 2024 & GPT-4 & None & INST, ZS & Yes & Training & Patient portal triage \\ 
\cite{jones2024teaching} & Text & EN & 2024 & Vicuna v1.1, Orca & IT & INST & No & Training & Report generation \\ 
\cite{alghamdi2024towards} & Text & AR & 2024 & ChatGPT & None & INST, RAG & Yes & Supplement & Health chatbot \\ 
\cite{wang2024twin} & Tabular & EN & 2024 & ChatGPT & None & FS, INST & No & Supplement & Clinical trial simulation \\ 
\cite{bucur2024leveraging} & Text & EN & 2024 & Llama, Alpaca & None & ZS, INST & No & Supplement & Depression symptom \\ 
\cite{woolsey2024utilizing} & Text & EN & 2024 & GPT-3.5/4 & None & ZS, INST & No & Supplement & Autism diagnosis \\ 
\cite{chen2025automated} & Text & EN & 2025 & ChatGPT, Llama, Gemini & None & INST & Request & Privacy & Stroke prediction \\ 
\cite{scroggins2025does} & Audio & EN & 2025 & GPT-4 & None & FS, CoT & No & Training & Dialogue analysis \\ 
\cite{albayrak2025enhancing} & Text & EN & 2025 & GPT-3.5 & None & ZS, INST & No & Supplement & Phenotype ontology \\ 
\cite{liu2025generating} & Text & EN & 2025 & GPT-3.5/4, Llama-2, Mistral & None & ZS, FS & No & Training & Radiology prediction \\ 
\cite{wang2025hypnos} & Text & ZH & 2025 & GPT-3.5, Claude & None & FS, RAG & Request & Training & Anesthesia QA \\ 
\cite{theodorou2025improving} & Tabular & - & 2025 & HALO & SFT & INST & Yes & Supplement & Patient outcome prediction \\ 
\cite{cai2025improving} & Text & ZH & 2025 & ChatGPT & None & FS, INST & Yes & Training & Mental health IE \\ 
\cite{barr2025large} & Tabular & EN & 2025 & GPT-4o & None & ZS, INST & Yes & Privacy & Data scarcity \\ 
\cite{chuang2025robust} & Text & EN & 2025 & GPT-3.5/4, Mistral & None & FS, INST & No & Privacy & Phenotype inference \\ 
\cite{kim2025small} & Text & EN & 2025 & GPT-4 & None & CoT & Request* & Training & Medical reasoning \\ 
\cite{li2025synthetic} & Text & ZH & 2025 & GPT-3.5 & None & INST & Request & Training & Database search \\ 
\cite{ashofteh2025targeted} & Text & EN & 2025 & Llama3-instruct & None & FS & No & Training & Metastases detection \\ 
\cite{ejiga2025text} & Image & EN & 2025 & Stable Diffusion & None & FS & Yes & Supplement & Colonoscopy   analysis \\ 
\cite{li2025topofm} & Image & EN & 2025 & ChatGPT API & None & INST & No & Supplement & Pathology analysis \\ 
\cite{vsuvalov2025using} & Text & ET & 2025 & GPT-2 & TAFT & ZS, FS & Request & Training & Clinical IE \\ 
\cite{miletic2025utility} & Tabular & EN & 2025 & DistilGPT-2 & SFT & INST & No & Training & Patient outcome prediction \\



\addlinespace[4pt]

\multicolumn{10}{@{}l@{}}{%
  \small
  \begin{minipage}{2\linewidth}
    \textit{Notes.} A dash (–) indicates unknown information. The ``Request*'' indicates data were stated to be publicly available but are either not yet released or temporarily inaccessible. Model names are reported as in the original papers; when only ``ChatGPT'' or ``ChatGPT API'' was mentioned, we list it accordingly.
  \end{minipage}
}\\

%

\end{tabular}
\end{adjustbox}

\end{figure}


\subsection{Synthetic Data Types and Medical Applications}

Increasing model developments for biomedical applications lead to the surging needs of synthetic data~\cite{ive2020generation, libbi2021generating}. 
In this section, we provide an overview of recent studies using synthetic data and address three critical questions: 
1) What types of synthetic data are being generated? 2) What modalities and clinical tasks do these datasets target? 3) Are those data resources accessible for reusable research, such as benchmarking and model training? 
To answer those questions, we examine several aspects of the collected studies and summarize them in Table~\ref{tab:generation_approach}, including data modality, data language, data size, published year, medical application, and data or model accessibility. 
We aim to provide insights into current and future trends of data types and biomedical applications.

Biomedical synthetic data has shown diverse medical topics, languages, and modalities. 
Most of the studies (55.9\%, $n=33$) are for general topics, while others include specific medical areas, such as suicide~\cite{spitale2023exploring,ghanadian2024socially}, colorectal cancer~\cite{ejiga2025text}, and radiology~\cite{liu2025generating}. 
For example, studies generate synthetic data for augmenting diagnosis accuracy for Post-traumatic stress disorder (PTSD)~\cite{wu2024callm} and Autism Spectrum Disorders (ASD)~\cite{woolsey2024utilizing}. 
English synthetic data accounted for the majority of generated datasets (84.7\%, $n=50$). 
Synthetic data were also generated in French~\cite{hiebel-etal-2023-synthetic}, Chinese~\cite{chen2024chatffa,wang2025hypnos,cai2025improving,li2025synthetic}, German~\cite{moser2024generating}, and Arabic~\cite{alghamdi2024towards}, demonstrating initial efforts to support non-English language resources. 
For example, Moser et al.~\cite{moser2024generating} employed multilingual LLMs to generate German emergency-medical dialogues simulating realistic interactions between ambulance staff and patients.
Most datasets focus on unstructured text modality (e.g., clinical narratives~\cite{wang2024notechat,chintagunta-etal-2021-medically,jones2024teaching} and discharge summaries~\cite{sufi2024addressing}), while a smaller but growing subset includes tabular~\cite{theodorou2023synthetic,ehrett2024leveraging,gao2024pressinpose,dobhal2024synthetic,wang2024twin,theodorou2025improving,barr2025large,miletic2025utility}, image~\cite{chen2024chatffa,ejiga2025text,li2025topofm}, and audio~\cite{scroggins2025does}. 
For example, Theodorou et al.~\cite{theodorou2023synthetic} generate novel tabular datasets of high-dimensional longitudinal EHR records to provide realistic, privacy-preserving alternatives for machine learning and statistical analysis, while Ejiga et al.~\cite{ejiga2025text} generate synthetic colonoscopy image datasets via fine-tuned text-to-image synthesis to augment training data for robust colorectal cancer detection and precise polyp segmentation. 
Data sizes vary across health issues (e.g., cancer and mental health), yet the majority has relatively small sets with fewer than 10K samples, indicating the critical utility of synthetic data under low-resourced scenarios.
For example, FairPlay~\cite{theodorou2025improving} generates synthetic data from authentic EHR records to bolster under-represented patient subgroups, boosting mortality‐prediction F1 scores by up to 21\% and markedly narrowing performance gaps across different demographic groups. 
The various synthetic datasets highlight a growing community to deploy the biomedical synthetic data.

Synthetic data generation in biomedical research increasingly supports a wide array of clinical applications. 
Those tasks cover diverse medical needs, including phenotype classification~\cite{zeinali2024symptom,woolsey2024utilizing,khademi2023data,li2023two,ashofteh2025targeted}, PTSD symptom extraction~\cite{wu2024callm}, de-identification (SynNote)~\cite{libbi2021generating}, clinical note summarization~\cite{sufi2024addressing,mishra2024synthetic}, and mental health assessment~\cite{bucur2024leveraging,bird2024generative,ghanadian2024socially}. 
For example, CALLM~\cite{wu2024callm} constructs clinical interview datasets to aid PTSD diagnosis by generating synthetic transcripts; and Barabadi et al.~\cite{ashofteh2025targeted} generated synthetic radiology reports to enable automatic detection of metastatic sites in cancer patients. 
Those applications broaden the utility of synthetic data for health challenges, such as data shortage~\cite{wu2024callm} and privacy concerns~\cite{libbi2021generating,sarkar2024identification,chuang2025robust,barr2025large,chen2025automated}. 
However, accessibility remains a central concern for advancing synthetic data utility (see Supplementary Table S1 for a detailed list of experimental datasets and their corresponding links).
Table~\ref{tab:generation_approach} shows 50.8\% ($n=30$) of the studies are explicitly described as accessible, providing open or partially open resources for reproducible development, while others’ accessibility details or licensing remain unclear.
Open datasets such as Syn-HPI~\cite{li2021synthetic}, ClinGen~\cite{xu-etal-2024-knowledge}, and Syn-EMS-Audio~\cite{weerasinghe2024real} serve as synthetic data benchmarks for evaluating new models and methods, while others—particularly those derived from sensitive clinical domains—may be restricted or anonymized to protect patient privacy. Ensuring clear, well-documented access protocols and licenses will be essential for fostering broader collaboration and translational impacts of synthetic data in biomedical research.

\subsection{Synthetic Data Generation Methodology}

\begin{figure}[!htbp]
\centering
\includegraphics[width=0.65\textwidth]{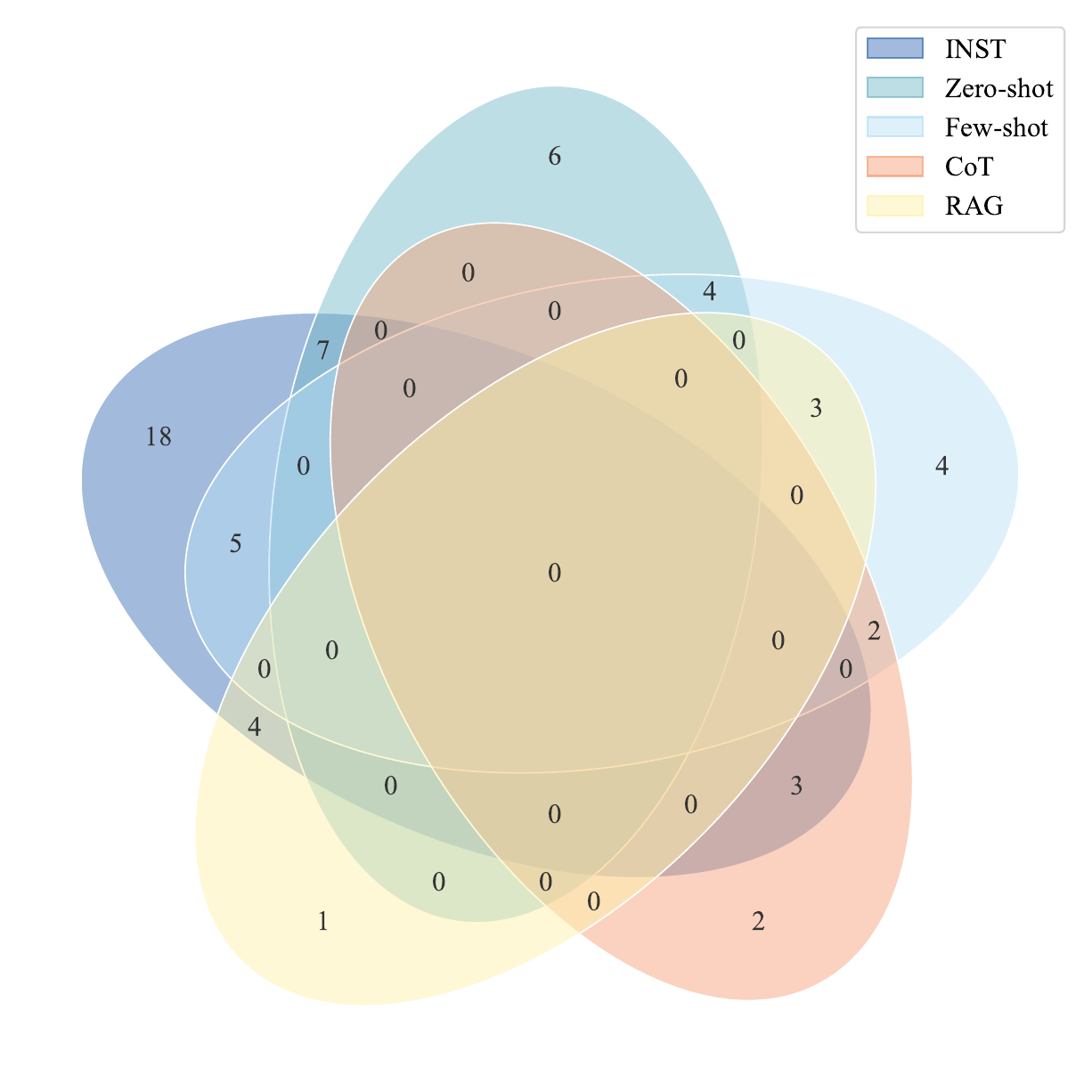}
\caption{Overlap among prompting strategies (INST, Zero-shot, Few-shot, CoT, RAG) in the reviewed studies. Each intersection represents studies combining multiple methods.}
\label{fig:venn5_prompting}
\end{figure}

Emerging Transformer-based large language models (LLMs) have significantly advanced synthetic data generation methodologies in biomedical research, enabling diverse applications, as summarized in Table~\ref{tab:generation_approach}. 
These novel approaches address challenges associated with limited clinical data by augmenting datasets, thus enhancing generalizability and mitigating overfitting in downstream machine learning tasks across various healthcare domains. 
In this section, we systematically examine generation methodologies among 59 reviewed studies, summarizing recent trends in model architectures, generation techniques, and integration strategies for synthetic data generation. 
Specifically, we aim to answer two key questions: 1) what predominant methods (e.g., prompt-based vs. specialized architectures) are currently employed for biomedical synthetic data generation; and 2) what trends and variations do exist regarding model selection (open-source vs. closed-source) and prompting strategies (zero-shot, few-shot, reasoning-augmented)? 
Because several studies employed combinations of approaches (e.g., both open- and closed-source models, or few-shot with CoT or RAG), each study was counted in all applicable categories.
In the case of prompting-based methods, studies employing multiple prompting strategies were summarized based on their two predominant prompting types for alignment, and the overlaps among these categories are visualized in Figure~\ref{fig:venn5_prompting}.

Our review identifies prompt-based generation as the predominant synthetic data generation method, employed by approximately 74.6\% ($n=44$) of the analyzed studies. 
Prompt-based approaches primarily rely on meticulously crafted textual instructions, leveraging fine-tuned LLMs, such as T5 and GPT variants (e.g., GPT-4) for diverse biomedical tasks~\cite{li2025synthetic,litake2024constructing,wang2024twin}. 
The most prevalent strategy is instruction-driven (INST) prompting, observed in 42.4\% ($n=25$) of studies, which guides generation through detailed, explicit instructions that typically encompass core task definitions, persona assignments, and specific constraints on the desired content, format, and style. 
It leverages the model's ability to follow complex commands without in-context examples, making it highly versatile. 
For instance, Sufi et al.\cite{sufi2024addressing} utilized detailed instructions to generate realistic discharge summaries, while Wang et al.\cite{wang2024taxonomy} employed structured prompts to synthesize patient portal messages for creating a corpus to analyze linguistic features and support downstream applications like patient portal message triage.
Zero-shot prompting represents another approach, used in 28.8\% ($n=17$) of studies, involving prompting LLMs without illustrative examples and relying solely on explicit task instructions and the pretrained knowledge encoded in LLMs~\cite{barr2025large}. 
This method is suitable for simpler tasks, such as dialogue summarization~\cite{chintagunta-etal-2021-medically} or generating structured medical reports~\cite{jones2024teaching}, where extensive contextual training is less critical. 
For example, Barr et al.~\cite{barr2025large} used zero-shot prompting with GPT-4o to synthesize perioperative clinical tables of patient demographics, surgical parameters and outcomes, demonstrating that the model can produce realistic synthetic data and preserve key statistical patterns based solely on qualitative instructions.
Few-shot prompting provides a small number of curated examples (typically 2-5) within the prompt to illustrate the task and guide LLMs for synthetic data generation, which accounts for 20.3\% ($n=12$) of studies. 
This approach uses limited curated examples to guide LLMs in generating clinically relevant synthetic data across diverse applications, including suicide symptom extractions~\cite{ghanadian2024socially}, phenotype classification~\cite{zeinali2024symptom,woolsey2024utilizing,khademi2023data,li2023two,ashofteh2025targeted}, and therapeutic dialogue generation~\cite{wang2024notechat,bird2024generative}. 
For example, Ghanadian et al.\cite{ghanadian2024socially} use a patient's suicidal narrative and a medication counseling dialogue as few-shot prompts to generate clinical texts for suicide symptom extraction and therapeutic dialogue generation, while Nievas et al.\cite{nievas2024distilling} use clinical trial inclusion and exclusion criteria paired with patient summaries as few-shot prompts to generate a synthetic patient–trial matching dataset.
Beyond providing instructions and examples, more sophisticated techniques aim to enhance the reasoning and factuality of generated data. 
Reasoning-augmented prompting, commonly known as Chain-of-Thought (CoT) and used in 10.2\% ($n=6$) of studies, enhances generation quality by instructing the model to first articulate its step-by-step reasoning process before producing the final output. 
This method improves the coherence and logical validity of synthetic data for complex clinical tasks. 
For example, Wu et al.~\cite{wu2024callm} applied CoT to generate synthetic PTSD diagnostic interviews by prompting the model to provide both a simulated patient's response and the clinical reasoning for it, ensuring the dialogue aligned with DSM-V criteria, while Kim et al.\cite{kim2025small} used CoT to improve medical reasoning for question-answering tasks. 
Furthermore, retrieval-augmented generation (RAG), improves factual accuracy by dynamically incorporating information from external knowledge bases into the prompt. 
This technique, found in 11.9\% ($n=7$) of studies, retrieves relevant documents (e.g., from clinical guidelines or medical literature) to ground the generation in verifiable, up-to-date information. 
For instance, Wang et al.\cite{wang2024dke} employed a RAG-like method for data augmentation, retrieving semantically similar synonyms from a knowledge graph to replace key terms. Similarly, Zafar et al.\cite{zafar2024ki-mag} enhanced medical QA by grounding generated answers with knowledge dynamically retrieved from the UMLS.

A notable trend is the model selection between open-source and closed-source LLMs, such as Llama vs GPT-4. 
Approximately 67.8\% ($n=40$) of the reviewed studies directly utilize closed-source models such as GPT-3.5 and GPT-4 through proprietary APIs. 
This approach prioritizes rapid development and output quality without extensive model customization. 
The remaining 32.2\% ($n=19$) of studies employ fine-tuning approaches on open-source models, which involves additional training of pretrained models on domain-specific datasets to adapt them for particular biomedical contexts.
Fine-tuning approaches can be categorized into several distinct types based on their adaptation strategies. 
Supervised fine-tuning (SFT) involves training models on labeled, task-specific datasets where input-output pairs are explicitly provided~\cite{ive2020generation,theodorou2025improving,miletic2025utility}. 
Ive et al.~\cite{ive2020generation} demonstrated this approach by fine-tuning Transformer models on real EHRs using a gap-filling task, where models learned to generate complete clinical documents conditioned on key phrases and clinical metadata for phenotype classification tasks.
Domain-adaptive fine-tuning (DAFT) adapts models to broad clinical domains by training on large corpora of domain-specific text without task-specific labels~\cite{libbi2021generating,hiebel-etal-2023-synthetic}. 
This approach helps models acquire clinical vocabulary, medical terminology, and domain-specific linguistic patterns. 
Libbi et al.~\cite{libbi2021generating} employed DAFT by fine-tuning an English GPT-2 model on Dutch clinical narratives with explicit PHI annotations, enabling joint generation of synthetic text and de-identification labels for training downstream NER models.
Task-adaptive fine-tuning (TAFT) focuses on specific downstream tasks by training models on task-relevant data~\cite{shakeri-etal-2020-end,li2021synthetic,lu2022textual,khademi2023data,theodorou2023synthetic,zafar2024ki-mag,vsuvalov2025using}. 
Lu et al.~\cite{lu2022textual} applied TAFT to GPT-2 by prepending class labels to MIMIC-III discharge summaries during training, enabling the model to generate synthetic clinical notes with embedded readmission status for downstream prediction tasks.
Instruction tuning (IT) trains models to follow structured commands and detailed instructions, enhancing their ability to generate content according to specific requirements~\cite{spitale2023exploring,jeong2024improving,delmas2024relation,jones2024teaching}. 
For instance, Jeong et al.~\cite{jeong2024improving} employed IT to fine-tune LLaMA2 models on medical instruction datasets, developing critic models that annotate and filter synthetic medical instructions to generate instruction sets for training generator models.
Fine-tuned open-source models demonstrate particular advantages in privacy-sensitive clinical contexts and specialized medical domains. 
However, these models typically require more computational resources and domain expertise compared to direct API usage of closed-source models. 
Studies comparing model performance indicate that while fine-tuned open-source models offer greater customization potential, they generally underperform closed-source models on standardized clinical reasoning benchmarks. 
For example, Jeong et al.~\cite{jeong2024improving} observed that fine-tuned LLaMA2 7B models achieved lower performance than GPT-4 on clinical reasoning tasks, despite domain-specific adaptation.

Approximately 25.4\% ($n=15$) of the reviewed studies fine-tuned pre-trained LLMs or explore specialized model architectures for synthetic data generation, 
including multimodal, multi-agent frameworks and custom-designed transformer architectures (e.g., HALO)~\cite{li2025topofm, ejiga2025text, weerasinghe2024real, scroggins2025does, gao2024pressinpose, wang2024notechat, theodorou2025improving, theodorou2023synthetic, ive2020generation}. 
Multimodal approaches integrate multiple data modalities, such as text, images, and audio, to create richer, more realistic synthetic datasets, such as colonoscopy image analysis~\cite{ejiga2025text}, digital pathology interpretation~\cite{li2025topofm}, and emergency medical services (EMS) assistance~\cite{weerasinghe2024real}. 
For example, CognitiveEMS~\cite{weerasinghe2024real} prompts an LLM with emergency protocols to generate synthetic cardiac-arrest dialogues, converts them to speech, and synchronizes the audio with AR smart-glasses video frames annotated by a zero-shot vision classifier, producing a unified dataset for cardiac emergency analysis. 
Multi-agent frameworks~\cite{alghamdi2024towards,wu2024callm,wang2024notechat} involve employing multiple interacting models or agents, each performing distinct roles, to collaboratively produce complex and realistic data, such as doctor-patient dialogues or clinical scenarios. 
For instance, NoteChat~\cite{wang2024notechat} employs a cooperative multi-agent framework with planning, role play, and refinement modules to generate synthetic patient–physician dialogues that mirror real clinical encounters. 
Custom transformer architectures\cite{theodorou2025improving,theodorou2023synthetic,ive2020generation} extend the standard decoder by integrating domain-specific metadata and employing hierarchical modeling strategies, first capturing broader structural information and then refining finer details, to generate structured and realistic synthetic clinical datasets tailored to specific medical scenarios. 
For example, Theodorou et al.~\cite{theodorou2023synthetic} demonstrate that their hierarchical autoregressive model generates privacy-preserving EHRs whose temporal and code distributions yield downstream prediction performance on par with real data. 
Collectively, these specialized frameworks enhance the fidelity and applicability of synthetic biomedical data for various complex health scenarios.

\begin{figure}[!htbp]
\captionsetup{type=table}
\centering
\caption{Evaluation characteristics of reviewed studies: disease, downstream task, data size, number of automated metrics (Metrics \#), inclusion of intrinsic evaluation (Intrinsic+), human involvement in generation or task evaluations (Human-in-the-Loop), and LLM-based evaluation (LLM Eval). 
CLS, NER, IE, RE, QA, NLI, and ASR refer to downstream tasks of classification, named entity recognition, information extraction, relation extraction, question answering, natural language inference, and automatic speech recognition, respectively.
}
\label{tab:evaluation}
\begin{adjustbox}{width=\textwidth, totalheight=0.77\textheight}
\begin{tabular}{cccccccc}
\textbf{Study} & \textbf{Disease} & \textbf{Task} & \textbf{Data Size (K)} & \textbf{Metrics \#} & \textbf{Intrinsic+} & \textbf{Human-in-the-Loop} & \textbf{LLM Eval} \\
\midrule
\cite{shakeri-etal-2020-end} & General & QA & 500,000 & 3 & Yes & No & No \\
\cite{ive2020generation} & Mental health & CLS & 11,000 & 6 & Yes & Yes & No \\
\cite{li2021synthetic} & General & NER & 500 & 4 & No & Yes & No \\
\cite{libbi2021generating} & General & NER & 355,000 & 5 & No & Yes & No \\
\cite{chintagunta-etal-2021-medically} & General & Summarization & 44800 & 3 & No & Yes & No \\
\cite{lu2022textual} & General & CLS & 48,393 & 3 & No & No & No \\
\cite{peng2023study} & General & RE, QA & 20B & 4 & No & Yes & No \\
\cite{hiebel-etal-2023-synthetic} & General & NER & 4,946 & 6 & Yes & Yes & No \\
\cite{khademi2023data} & Febrile convulsions & CLS & 4,917 & 3 & No & Yes & No \\
\cite{spitale2023exploring} & Suicide & CLS & 50 & 2 & No & Yes & No \\
\cite{theodorou2023synthetic} & General & CLS & 975,788 & 8 & No & Yes & No \\
\cite{li2023two} & Alzheimer & CLS & 32,116 & 4 & No & Yes & No \\
\cite{sufi2024addressing} & General & CLS & 70 & 3 & No & Yes & No \\
\cite{wu2024callm} & PTSD & CLS & 1,900 & 3 & No & Yes & No \\
\cite{chen2024chatffa} & Fundus Fluorescein Angiography & QA & 4,110,581 & 5 & No & Yes & No \\
\cite{litake2024constructing} & Acute Renal Failure & CLS & 9,000 & 7 & No & Yes & No \\
\cite{sarkar2024identification} & General & CLS & 4,033 & 5 & No & No & No \\
\cite{nievas2024distilling} & General & CLS & 2,000 & 8 & No & Yes & No \\
\cite{wang2024dke} & General & NLI & - & 5 & No & No & No \\
\cite{moser2024generating} & General & - & 200 & 1 & Yes & No & No \\
\cite{bird2024generative} & Psychological Therapy & - & 3,508 & 9 & Yes & No & No \\
\cite{jeong2024improving} & General & QA & 18,854 & 3 & No & No & No \\
\cite{zafar2024ki-mag} & General & QA & 30,172 & 8 & No & Yes & No \\
\cite{xu-etal-2024-knowledge} & General & CLS, RE, NER & 5,000 & 7 & Yes & No & No \\
\cite{guevara2024large} & General & IE, CLS & 1,800 & 3 & No & Yes & No \\
\cite{ehrett2024leveraging} & COVID-19 & CLS & 2,050 & 1 & No & Yes & No \\
\cite{wang2024notechat} & General & Dialogue Generation & 30,000 & 7 & Yes & Yes & Yes \\
\cite{gabriel2024development} & General & CLS & 592 & 7 & No & No & No \\
\cite{gao2024pressinpose} & Gait-based Disease & Pose Estimation & 14,400 & 4 & Yes & No & No \\
\cite{kweon2024publicly} & General & IE & 158,000 & 1 & Yes & Yes & Yes \\
\cite{weerasinghe2024real} & General & ASR & 59-min & 2 & No & No & No \\
\cite{delmas2024relation} & General & NER, RE & 25,239 & 1 & No & No & No \\
\cite{ghanadian2024socially} & Suicide & CLS & 4,897 & 2 & No & Yes & No \\
\cite{zeinali2024symptom} & Cancer & CLS & 180 & 2 & No & No & No \\
\cite{mishra2024synthetic} & General & Summarization & 5,256 & 4 & Yes & Yes & Yes \\
\cite{zhang-etal-2024-synthetic} & General & QA & 147,980 & 1 & No & No & No \\
\cite{dobhal2024synthetic} & Skeleton pose & CLS & 5,400 & 2 & Yes & No & No \\
\cite{wang2024taxonomy} & General & - & 450 & 2 & Yes & Yes & No \\
\cite{jones2024teaching} & General & Summarization & 100,000 & 6 & No & No & Yes \\
\cite{alghamdi2024towards} & General & QA & 150 & 4 & No & Yes & Yes \\
\cite{wang2024twin} & Breast cancer & CLS & - & 6 & Yes & No & No \\
\cite{bucur2024leveraging} & Depression & Semantic Search & 2,700 & 4 & No & No & No \\
\cite{woolsey2024utilizing} & Autism & CLS & 4,200 & 3 & No & Yes & No \\
\cite{chen2025automated} & Stroke thrombolysis contraindications & CLS & 150 & 6 & No & Yes & No \\
\cite{scroggins2025does} & General & CLS & 3,042 & 4 & No & Yes & No \\
\cite{albayrak2025enhancing} & Phenotype ontology & IE & 458,574 & 3 & No & No & No \\
\cite{liu2025generating} & Limb Fractures & CLS & 4,600 & 3 & Yes & No & Yes \\
\cite{wang2025hypnos} & Anesthesiology & QA & 180,000 & 5 & No & Yes & Yes \\
\cite{theodorou2025improving} & General & CLS & - & 2 & No & No & No \\
\cite{cai2025improving} & Mental health & IE & 5,802 & 2 & No & Yes & No \\
\cite{barr2025large} & Perioperative clinical data & - & 6,166 & 1 & Yes & No & No \\
\cite{chuang2025robust} & General & CLS & 88,353 & 3 & Yes & No & No \\
\cite{kim2025small} & General & QA & 77,776 & 5 & No & Yes & Yes \\
\cite{li2025synthetic} & General & CLS & 30,000 & 2 & No & No & No \\
\cite{ashofteh2025targeted} & Cancer & CLS & 74,450 & 1 & No & Yes & No \\
\cite{ejiga2025text} & Colorectal cancer & CLS, Segmentation & 1,800 & 7 & No & No & No \\
\cite{li2025topofm} & Breast cancer & CLS, Segmentation & - & 4 & No & Yes & No \\
\cite{vsuvalov2025using} & General & NER & 4,100 & 3 & No & Yes & No \\
\cite{miletic2025utility} & Breast Cancer, Diabetes & CLS & 1,165,660 & 3 & No & No & No
\end{tabular}
\end{adjustbox}
\end{figure}

\subsection{Quality Assessment and Evaluation Metrics}

Ensuring the quality of the synthetic data is the key to building precise models and medical applications. In this section, we assess our collected studies from the synthetic data assessment and evaluation perspective, as shown in Table~\ref{tab:evaluation}. 
We aim to answer the following three questions: 1) What metrics and evaluation approaches are commonly used for assessing synthetic data quality? 2) What are the major challenges in model evaluations using synthetic data across biomedical diseases and tasks; and 3) What are the emerging trends in synthetic data quality control for biomedical research? 
To answer those questions, we systematically examined the collected 59 studies by disease topics, types of evaluation metrics (automated and human approaches), downstream biomedical tasks, and emerging strategies (e.g., LLM-based and intrinsic evaluations). 
We aim to highlight current practices, identify persistent challenges, and discuss notable shifts toward more robust and clinically meaningful data and model assessments in the biomedical synthetic data generation.

We examined two major evaluation metrics across diverse biomedical applications, intrinsic and extrinsic metrics, which are the essential measurements to select synthetic data and computational models. 
Intrinsic metrics (e.g., perplexity~\cite{kweon2024publicly} or similarity measures such as KL Divergence~\cite{liu2025generating}) assess properties of the synthetic data independently from downstream tasks, while extrinsic metrics rely on downstream tasks, such as classification~\cite{ive2020generation}. 
As shown in Table~\ref{tab:evaluation}, we can observe that the intrinsic metrics are not widely adopted and only count 27.1\% ($n=16$) of the studies. 
For example, Study~\cite{xu-etal-2024-knowledge} and Study~\cite{mishra2024synthetic} employ intrinsic metrics to directly compare the statistical distribution of real and synthetic datasets, while most works, such as Studies~\cite{lu2022textual,khademi2023data}, and~\cite{albayrak2025enhancing}, rely primarily on extrinsic evaluation through classification (CLS) or information extraction (IE) tasks. 
Commonly used metrics include accuracy~\cite{spitale2023exploring}, F1 score~\cite{khademi2023data}, BLEU~\cite{chen2024chatffa}, and other task-specific measurements, such as ROUGE~\cite{kim2025small} and AUCROC~\cite{lu2022textual}. 
The studies cover heterogeneous downstream tasks, such as de-identification for privacy protection~\cite{sarkar2024identification}, phenotype prediction~\cite{albayrak2025enhancing}, disease entity extraction~\cite{kweon2024publicly}, and clinical note summarization~\cite{mishra2024synthetic,chintagunta-etal-2021-medically}. 
Privacy protection assessment was reported in 20.3\% ($n=12$) of the studies, primarily through membership and attribute inference attacks~\cite{sarkar2024identification}, adversary test~\cite{theodorou2023synthetic}, and memorization evaluation~\cite{ive2020generation}. 
Those studies cover multiple medical domains, with most studies focusing on general biomedical applications ($n=33$) and others targeting 25 specific medical conditions and health-related topics, such as Alzheimer’s~\cite{li2023two}, breast cancer~\cite{wang2024twin}, and depression~\cite{bucur2024leveraging}.

\begin{figure*}[ht]
\centering
\includegraphics[width=0.7\textwidth]{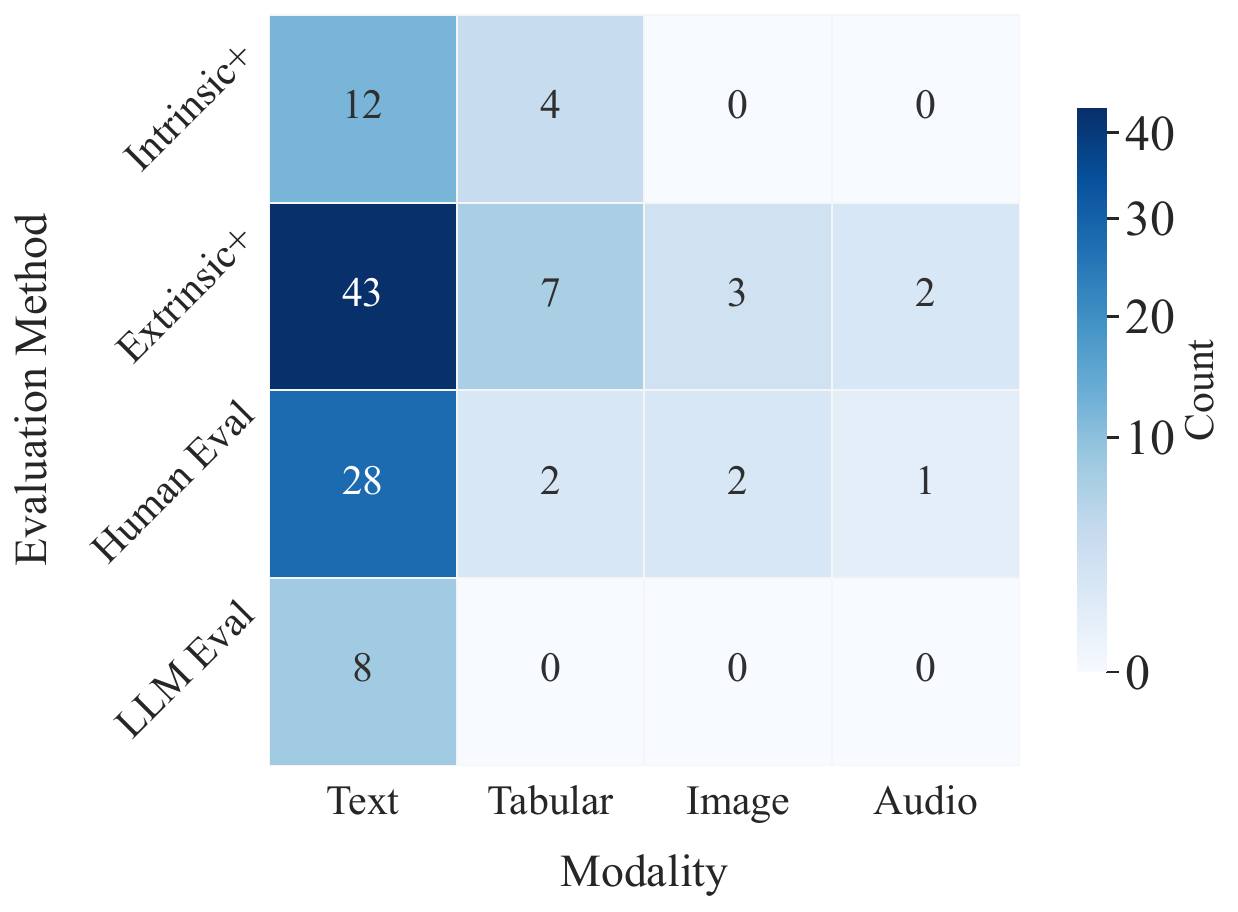}
\caption{Heatmap linking data modality to evaluation method, summarizing the number of reviewed studies across four data modalities (Text, Tabular, Image, Audio) and four evaluation types: intrinsic evaluation (Intrinsic+), extrinsic evaluation (Extrinsic+), human evaluation (Human Eval; as in ``Human-in-the-Loop'' in Table~\ref{tab:evaluation}), and LLM-based evaluation (LLM Eval).}
\label{fig:Modality_eval_heatmap}
\end{figure*}

Beyond automated metrics, human evaluation has played a selective yet important role in the assessment of synthetic biomedical data or its utility in different tasks (e.g., phenotype inference or clinical note summary), particularly for tasks where clinical relevance or nuanced interpretation is required. Human evaluation~\cite{nievas2024distilling, li2023two, wang2024notechat} typically involves domain experts to determine if synthetic data are meaningful, such as clinicians or biomedical researchers, who assess the synthetic data for factors like clinical validity. 
Human evaluation was included in a minority of studies (44.1\%). 
Those studies cover tasks such as clinical entity recognition~\cite{li2021synthetic}, clinical note summarization~\cite{chintagunta-etal-2021-medically, mishra2024synthetic}, or question answering~\cite{zafar2024ki-mag}, where expert judgment adds valuable context beyond quantitative scores. 
More recently, a few studies~\cite{wang2024notechat,kweon2024publicly,mishra2024synthetic,jones2024teaching} (13.6\% of our studies) explore using large language models (LLMs) as automated evaluators and leverage their ability to perform nuanced judgments, which approximates human assessment. 
Common LLM-based evaluation metrics employed in these studies include factual consistency~\cite{jones2024teaching}, clinical correctness~\cite{kweon2024publicly}, and error identification~\cite{alghamdi2024towards}, often measured through prompt-based scoring or rubric-guided assessments conducted by the LLM itself. 
For example, a recent study~\cite{jones2024teaching} prompts GPT-4 if the context supports / contradicts the response, or if there’s not enough information as a hallucination evaluation, whereas another study~\cite{kweon2024publicly} uses a rubric-guided prompt, requesting GPT-4 to score each response on a four-level scale: Unacceptable, Poor, Satisfactory, and Excellent.
However, LLM-based evaluations remain an emerging approach and have not yet replaced the need for statistical metrics or domain expert involvement, especially in clinically sensitive or complex scenarios~\cite{chen2025evaluating}.
The overall distribution of evaluation methods across data modalities is summarized in Figure~\ref{fig:Modality_eval_heatmap}. 
As shown, text-based studies dominate all evaluation types, while non-text modalities such as tabular, image, and audio remain underrepresented, particularly in intrinsic and LLM-based evaluations.

Despite substantial progress, several challenges persist in the quality control of biomedical synthetic data. 
A key issue is the lack of standardized evaluation frameworks, resulting in wide variability in both the selection and reporting of metrics across studies and disease domains~\cite{xu-etal-2024-knowledge}. 
This heterogeneity complicates direct comparison between methods and limits the transferability of findings across medical applications and diseases. 
Additionally, Table~\ref{tab:evaluation} shows that studies still underutilize intrinsic evaluation, focusing instead on extrinsic, task-specific benchmarks that may not fully capture the utility or the limitations of synthetic data. 
Emerging trends suggest a gradual shift toward more comprehensive assessments, including extrinsic and intrinsic evaluation integrations, LLM-based metric adoption, and human professional engagements.

\section{Methods}\label{Methods}

\begin{figure*}[ht]
\centering
\includegraphics[width=0.83\textwidth]{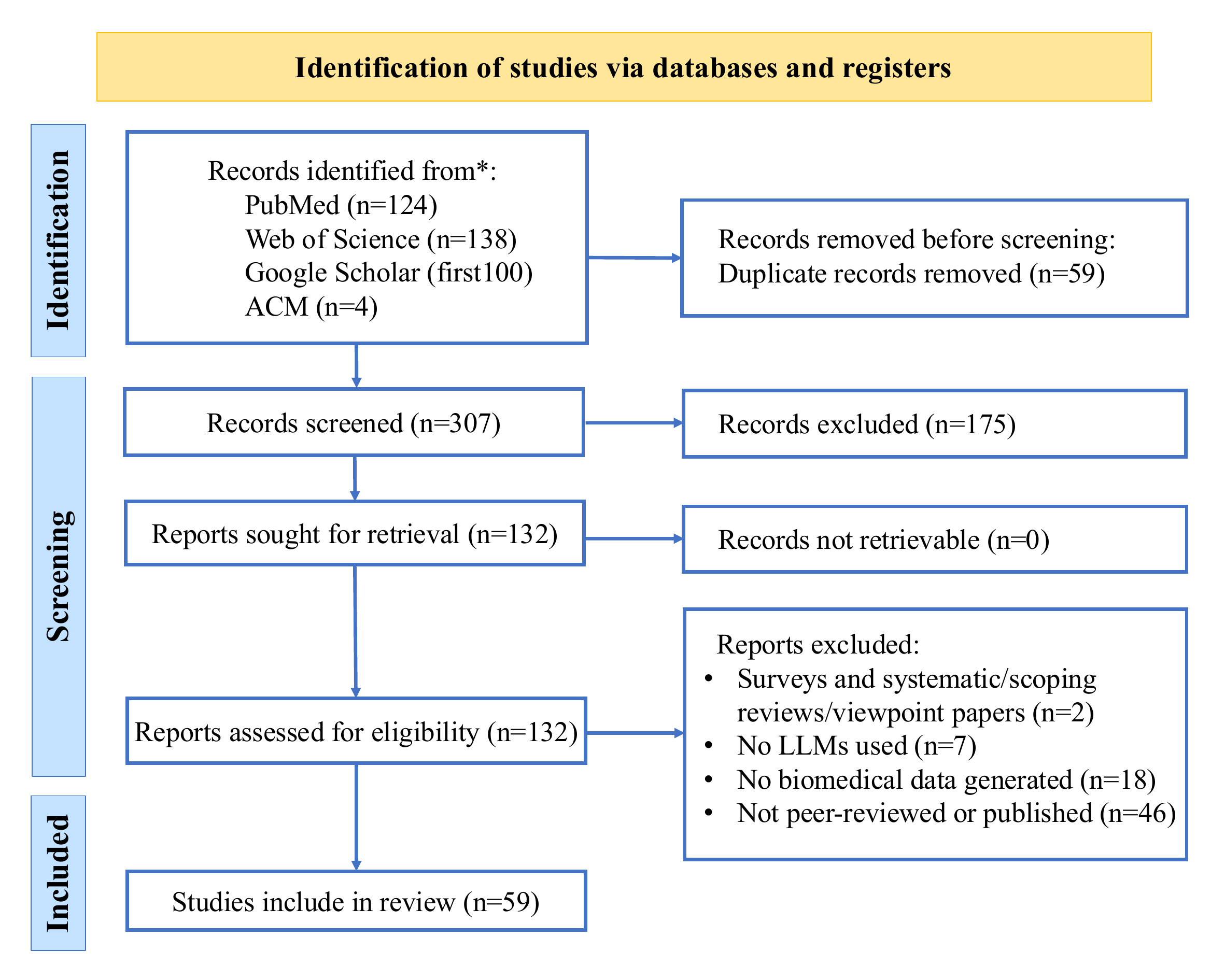}
\caption{PRISMA-ScR flow diagram.}
\label{fig:PRISMA-ScR_flow}
\end{figure*}

\paragraph{Data Source and Search} We conduct searches between June and August 2024 with an additional search in May 2025 following the Preferred Reporting Items for Systematic Reviews and Meta-Analyses (PRISMA) guidelines~\cite{Tricco2018PRISMA} and the PRISMA Extension for Scoping Reviews (PRISMA-ScR)~\cite{PRISMA-ScR2018} (Figure \ref{fig:PRISMA-ScR_flow}). 
Recognizing the rapid developments of Large Language Models (LLMs) in the biomedical field, our review encompassed peer-reviewed articles between January 1, 2020, and April 5, 2025 (inclusive) from multiple database sources, including ACM Digital Library, PubMed, Web of Science, and Google Scholar. 
Supplemental articles were gathered by reviewing article bibliographies and by soliciting suggestions from XH, HW, IH, and ZH. 
Our search strategy utilized a logical combination of keywords, and the full boolean search string used for Web of Science was:
\texttt{TS=(``large language model*'' OR ``LLM*'') AND TS=(synthetic*) AND TS=(data*) AND TS=(biomedical* OR medical* OR clinical* OR health*).} 
Search strings for other databases followed a similar logic and are provided in the Appendix~\ref{appendix:db-search-strategy}.
We primarily limited the search to titles and abstracts, expanding to full texts where this function was unavailable. For Google Scholar, we conducted a keyword search, sorted by relevance, and selected the first 100 studies. 
The search strategies for each database were initially formulated in the early stages of the study and refined through team discussions and preliminary analysis of the results. 
The final database search was conducted on April 5, 2025, covering all publications available by that date.

\paragraph{Study Selection} 
All records retrieved from the database searches were imported into Mendeley for reference management. 
Duplicates were removed in Mendeley based on matching titles, authors, and publication details before screening.
Screening and eligibility assessment were conducted collaboratively using a shared Excel spreadsheet. 
Titles and abstracts were initially screened against predefined eligibility criteria by at least two independent reviewers, excluding records that clearly did not meet inclusion criteria and retaining ambiguous ones for full-text screening. 
Our inclusion criteria include: 1) an article conducts biomedical research; 2) an article uses Large Language Models (LLMs) for synthetic generation of biomedical data; and 3) an article has undergone peer review and been published in a journal or in conference proceedings within the range of 01/01/2020 and 04/05/2025. 
Our exclusion criteria include: 1) an article is a survey, literature review, news article, editorial, letter, opinion, study protocol, and comment; 2) an article is not in English; 3) articles do not study biomedical large language models; and 4) an article is not peer-reviewed and published in a journal or conference proceedings. 
Three undergraduate research assistants in the computer science major from the XH lab, who had significant experience and trainings in biomedical studies. 
They assisted HS and XH in screening candidate articles. 
Each record was independently reviewed by at least two individuals, with decisions documented in a shared spreadsheet. 
Discrepancies were discussed to reach consensus, and unresolved cases were adjudicated by XH or HW.


\paragraph{Data Extraction} 
Selected articles were mainly processed by two individuals (HR and XH), who are experts in natural language processing (NLP) and biomedical informatics. 
They independently reviewed the full text of each selected article and extracted key metadata using a predefined data extraction template developed in Microsoft Excel.
The template was pilot-tested on five randomly selected studies to ensure that all relevant information could be consistently captured and refined before full data extraction.
The extracted data focused on three main aspects: 
1) Synthetic data types and applications, including data characteristics , such as language and domain; 
2) Synthetic data generation methods, presenting generation pipelines (models used to generate synthetic data) and their data modalities (e.g., text, image, audio); 
3) Quality assessment and evaluation metrics summarizing medical applications, evaluation approaches, and evaluation settings. 
Each field in the template corresponded to one column in the summary tables (e.g., Table~\ref{tab:generation_approach} and Table~\ref{tab:evaluation}), allowing for transparent and reproducible aggregation of results.
Discrepancies in data extraction were resolved through discussion between HR and XH, with further arbitration by XH, HW, and IH when necessary. 
All extraction processes were documented for replication purposes.


\section{Discussion}\label{Discussion}


Our scoping review has presented the current developments and applications of biomedical synthetic data generation, which is being facilitated by large language models (LLMs).
In this section, we highlight several insightful takeaways of emerging methodological and evaluative trends and identify promising areas for future biomedical research and applications using synthetic data generation. 


This review shows a clear shift towards leveraging synthetic data to overcome biomedical research challenges, such as data scarcity, privacy restrictions, and diverse clinical conditions. 
While synthetic data has been used effectively across varied clinical applications (e.g., phenotype ontology extraction~\cite{albayrak2025enhancing} and cancer diagnosis~\cite{ejiga2025text,li2025topofm,zeinali2024symptom}), the scale still varies across data types and disease fields, reflecting the nuanced capabilities and limitations of current LLMs. 
To navigate these complexities and enhance data quality, our review identifies a strategic adoption of advanced generation methodologies. 
Fine-tuning approaches, such as task-adaptive fine-tuning (TAFT), are employed to specialize models for specific clinical contexts, thereby improving domain relevance~\cite{shakeri-etal-2020-end,li2021synthetic,lu2022textual,khademi2023data}.
In parallel, sophisticated prompting techniques address concerns of logical coherence and factuality. For instance, reasoning-augmented methods like Chain-of-Thought (CoT) improve the clinical plausibility of outputs~\cite{wu2024callm,kim2025small}, while knowledge-augmented generation (RAG) grounds synthetic data in verifiable external sources to mitigate hallucination~\cite{wang2024dke,zafar2024ki-mag}. 
These methods are applied to various data types, with clinical narratives and unstructured texts remaining the most common data source, due primarily to their widespread availability and the inherent strengths of LLMs in natural language processing tasks.
We can also find an emerging trend toward multimodal and structured synthetic EHR, which indicates increasing expansions across more clinical settings, such as imaging, audio, and wearable sensors~\cite{gao2024pressinpose}.

However, the advances indicate potential concerns in quality augmentation and assessments. 
Our review shows that current studies are significantly depending on prompting closed-source LLMs, like GPT-4. 
While few-shot and zero-shot approaches may achieve some successful cases~\cite{wu2024callm, delmas2024relation, litake2024constructing}, prompting closed-source LLMs (e.g., GPT-4) prioritizes rapid development and sophisticated output quality, though their proprietary nature poses challenges for transparency, reproducibility, and regulatory approval in clinical settings.
This reliance on proprietary LLMs presents several challenges. 
Access to proprietary models may be restricted by cost, availability, or data governance policies. 
For example, MIMIC-IV (Medical Information Mart for Intensive Care)~\cite{johnson2023mimic2.2} does not allow for uploading data to those closed-source LLMs.
Reproducibility faces additional barriers from the non-deterministic nature of these APIs and inconsistent reporting of prompts or decoding settings. 
Additionally, among the 43 studies that used prompt-based generation, many did not report essential details such as model versions, complete prompts used, or decoding parameters (e.g. sampling temperature), making replication difficult~\cite{sufi2024addressing,li2025topofm}.
Hallucination and fabrication risks often lack systematic evaluation, potentially leading to errors in downstream applications. 
Many studies rely on n-gram-based metrics such as ROUGE, which assess textual fluency but cannot verify clinical accuracy. 
More comprehensive factuality assessments, such as Hallucination Rate quantification or ontology-based checks like UMLS-F1~\cite{mishra2024synthetic}, were adopted by only a minority of studies, leaving a gap in ensuring the factual accuracy and clinical safety of synthetic biomedical data.
Addressing these challenges requires implementing basic safeguards, such as disclosing generation details (e.g., prompts, model versions, decoding parameters), avoiding same-family generator–evaluator configurations, and conducting targeted expert reviews to assess factual accuracy.
Future work should focus on developing and benchmarking open-source alternatives and exploring methods that reduce reliance on proprietary systems.

Our review indicates future research in biomedical synthetic data generation may evolve along three dimensions.
There is a clear trend towards generating multimodal synthetic datasets to simulate more real-world and complex clinical scenarios, such as radiology~\cite{liu2025generating,ashofteh2025targeted}, pathology~\cite{li2025topofm}, and emergency care~\cite{moser2024generating}. 
Future synthetic data generation may depend on more precise LLM-based systems by knowledge-guided generations~\cite{xu-etal-2024-knowledge}, multi-agent frameworks~\cite{wang2024notechat}, and human-LLM collaborations~\cite{ehrett2024leveraging}, which leverage clinical knowledge to enhance factuality and specificity of synthetic data qualities. 
In terms of evaluation, more standardized, multi-dimensional, and human-centered assessments are critical.
Future studies on biomedical synthetic data generation may consider a mixture of intrinsic and extrinsic metrics, human-in-the-loop approaches, and task-agnostic validations to ensure fidelity, utility, and privacy of synthetic data.
We envision the directions will be essential to advance biomedical research and applications by the synthetic data generations and LLMs.

\paragraph{Limitations} 
This scoping review should be interpreted in light of several constraints. 
First, the search was restricted to four major databases (ACM Digital Library, PubMed, Web of Science, and Google Scholar) and to peer-reviewed venues; grey literature, non-indexed conference proceedings, and pre-prints may therefore be under-represented. 
Second, we limited inclusion to studies published in English between 1 January 2020 and 5 April 2025, so potentially relevant work in other languages or outside this time frame was not captured. 
Third, consistent with the PRISMA-ScR framework, we performed no critical appraisal of individual study quality; consequently, the methodological robustness of the included papers was not formally assessed. 
Additionally, the heterogeneity of study designs, synthetic-data tasks, evaluation metrics, and biomedical application domains precluded quantitative synthesis and limits the generalizability of pooled observations. 
Reproducibility is further limited by reliance on proprietary, non-deterministic APIs and incomplete reporting of prompts/decoding settings; hallucination audits were inconsistently performed across included studies. 
Lastly, given the rapid pace of innovation in LLM-based synthetic data generation, newly published methods appearing after our search window may not be reflected here, underscoring the need for periodic updates of this evidence map.

\section{Conclusion}\label{Conclusion}



Our scoping review examines the current trends and challenges of utilizing synthetic data generated by language models in biomedical fields with a focus on utility and quality aspects.
The results indicate that to fully utilize the LLM-generated synthetic data, the field may address several critical issues, including standardized evaluation frameworks, increased transparency, human-AI collaboration, and accessible models and data.
Although our review was limited by the rapid pace of advancements and a primary focus on English-language publications, these challenges present important opportunities for collaboration and innovation. 
By promoting reproducibility, rigorous evaluation, and expert feedback for the synthetic generation approach, our study shows the integration of high-quality synthetic data can benefit biomedical research and clinical practice.

\backmatter







\section*{Statements and Declarations}



\subsection*{Funding}
HR, WL, and XH were partially supported by the National Science Foundation under Award IIS-2245920 and IIS-2440381, and IH was partially supported by National Cancer Institute Award R01CA258193. HW was supported by NIH Office of Strategic Coordination (Common Fund) R03OD038389. ZH was supported by the Agency for Healthcare Research and Quality Award R21HS029969, the National Library of Medicine Award R21LM013911, the National Institute of Mental Health Award R21MH137736, and the National Institute on Aging Award R01AG064529. 

\subsection*{Conflict of interest}
The author(s) declared no potential conflicts of interest with respect to the research, authorship, and/or publication of this article.

\subsection*{Ethics approval and consent to participate}
Not applicable.

\subsection*{Consent for publication}
Not applicable.

\subsection*{Data availability} 
Data are available upon request to the corresponding author.

\subsection*{Materials availability}
Not applicable.

\subsection*{Code availability}
Not applicable.

\subsection*{Author contribution}  
H.R. and W.L. led the literature search, data extraction, and formal analysis. H.R. and W.L. made equal contributions to this study. H.W. contributed to initial data collection, methodology development, and manuscript preparation. W.L. joined the process at a later stage. IC.H. and Z.H. provided domain expertise, supervised the review process, and revised the manuscript. X.H. conceived and designed the study, coordinated the project, and contributed to writing, editing, and finalizing the manuscript. All authors reviewed, edited, and approved the final manuscript.

\clearpage
\begin{appendices}

\section{Database-specific search strategies}\label{appendix:db-search-strategy}

\begin{table}[ht]
\centering
\begin{tabularx}{\textwidth}{l >{\RaggedRight}X}
\toprule
\textbf{Database} & \textbf{Search string} \\
\midrule
PubMed &
(``large language model*''[Title/Abstract] OR ``LLM*''[Title/Abstract]) AND (synthetic*[Title/Abstract]) AND (data*[Title/Abstract]) AND (biomedical*[Title/Abstract] OR medical*[Title/Abstract] OR clinical*[Title/Abstract] OR health*[Title/Abstract]) \\
\addlinespace
ACM Digital Library &
Abstract:(``large language model*'' OR ``LLM*'') AND Abstract:synthetic* AND Abstract:data* AND Abstract:(biomedical* OR medical* OR clinical* OR health*) \\
\addlinespace
Web of Science &
TS=(``large language model*'' OR ``LLM*'') AND TS=(synthetic*) AND TS=(data*) AND TS=(biomedical* OR medical* OR clinical* OR health*) \\
\addlinespace
Google Scholar &
(``large language model'' OR ``large language models'' OR LLM) AND synthetic AND data AND (biomedical OR medical OR clinical OR health) \\
\bottomrule
\end{tabularx}
\caption{Search strategies for each database included in this review.}
\label{tab:db_search_strings_optimized}
\end{table}




\end{appendices}

\bibliography{syn-data-gen}

\end{document}